\documentclass[10pt,journal,onecolumn]{IEEEtran}
\usepackage{times}  
\usepackage{helvet}  
\usepackage{courier}  
\usepackage{url}  
\usepackage{graphicx}  
\usepackage{latexsym}
\usepackage{CJKutf8}
\usepackage{multirow}
\usepackage{tabularx}
\usepackage{caption}
\usepackage{subcaption}
\usepackage{amsmath}
\usepackage{color}

\newcommand{\furl}[1]{\footnote{\url{http://#1}}}

\begin{document}

\title{Phonetic-enriched Text Representation\\for Chinese Sentiment Analysis\\with Reinforcement Learning}

\author{Haiyun Peng,
        Yukun Ma,
        Soujanya Poria,
        Yang Li,
        Erik Cambria
\IEEEcompsocitemizethanks{\IEEEcompsocthanksitem All authors are from the School of Computer Science and Engineering at Nanyang Technological University
\IEEEcompsocthanksitem E-mails: see http://sentic.net/team}}

\markboth{IEEE Transactions on Affective Computing}%
{Peng \MakeLowercase{\textit{et al.}}}

\IEEEtitleabstractindextext{%
\begin{abstract}
The Chinese pronunciation system offers two characteristics that distinguish it from other languages: deep phonemic orthography and intonation variations. We are the first to argue that these two important properties can play a major role in Chinese sentiment analysis. Particularly, we propose two effective features to encode phonetic information. Next, we develop a Disambiguate Intonation for Sentiment Analysis (DISA) network using a reinforcement network. It functions as disambiguating intonations for each Chinese character (pinyin). Thus, a precise phonetic representation of Chinese is learned. Furthermore, we also fuse phonetic features with textual and visual features in order to mimic the way humans read and understand Chinese text. Experimental results on five different Chinese sentiment analysis datasets show that the inclusion of phonetic features significantly and consistently improves the performance of textual and visual representations and outshines the state-of-the-art Chinese character level representations. 
\end{abstract}

}

\maketitle

\section{Introduction}\label{sec:introduction}
In recent years, sentiment analysis has become increasingly popular for processing social media data on online communities, blogs, wikis, microblogging platforms, and other online collaborative media~\cite{campra}. Sentiment analysis is a branch of affective computing research~\cite{porrev,zadeh2018multi} that aims to classify text -- but sometimes also audio and video~\cite{pordep,majumder2018dialoguernn} -- into either positive or negative -- but sometimes also neutral~\cite{chadis}. Sentiment analysis techniques can be broadly categorized into symbolic and sub-symbolic approaches: the former include the use of lexicons~\cite{banlex}, ontologies~\cite{draont}, and semantic networks~\cite{camnt5} to encode the polarity associated with words and multiword expressions; the latter consist of supervised~\cite{onesta}, semi-supervised~\cite{hussem} and unsupervised~\cite{liilea} machine learning techniques that perform sentiment classification based on word co-occurrence frequencies. There are also a few hybrid approaches~\cite{camsen,poria2013music} that leverage an ensemble of symbolic and sub-symbolic techniques for polarity detection.

While most literature addresses the problem in a language-independent approach, Chinese sentiment analysis in fact requires tackling language-dependent challenges due to its unique nature, including word segmentation~\cite{huang2007chinese}, compositional analysis~\cite{sun2014radical,li2015component,shi2radical,yin2016multi,penrad}. There are two main characteristics distinguishing Chinese from other languages. Firstly, it is a \emph{pictogram} language~\cite{hansen1993chinese}, which means that symbols (called Hanzi) intrinsically carry meanings. Multiple symbols might form a new single symbol via geometric composition.The hieroglyphic nature of Chinese writing system differs from many Indo-European languages such as English or German. It has therefore inspired many works to explore the sub-word components (such as Chinese character and Chinese radicals) via a textual approach~\cite{yin2016multi,chen2015joint,sun2014radical,li2015component,shi2radical,penrad}. 
The other research line models the compositionality 
using the visual presence of the characters~\cite{su2017learning,liu2017learning} by the means of extracting visual features from bitmaps of Chinese characters to further improve the Chinese textual word embeddings.

The second characteristic of Chinese is that it is a language of deep phonemic orthography according to the orthographic depth hypothesis~\cite{frost1987strategies,katz1992reading}. In other words, it is hard to support the recognition of words involving the language phonology.  Each symbol of modern Chinese language can be phonetically transcribed into a romanized form, called PinYin, consisting of an initial (optional), a final, and the tone. More specifically, as a tonal language, one single syllable in modern Chinese can be pronounce with five different tones, i.e., 4 main tones and 1 neutral tone (shown later in Table~\ref{act_tone}). We would argue that this particular form of Chinese language provides semantic cues complementary to its textual form as illustrated in Table~\ref{into-tab}. Despite its important role in Chinese language, to the best of our knowledge, it has not yet been explored by existing work for NLP tasks of Chinese Language. 

In this work, we argue that the second factor of Chinese language can play a vital role in Chinese natural language processing especially sentiment analysis. Particularly, to consider the deep phonemic orthography and intonation variety of the Chinese language, we propose two steps to learn Chinese phonetic information.

\begin{table}[h]
	\centering
	\caption{Examples of intonations that alter meaning and sentiment.}
	\label{into-tab}
	\resizebox{0.5\textwidth}{!}{%
\begin{tabular}{cccc}
	\hline
	Text               & Pronunciation & Meaning   & Sentiment polarity \\ \hline
	\multirow{2}{*}{\begin{CJK*}{UTF8}{gbsn}空\end{CJK*}} & k\=ong            & Empty     & Neutral             \\
	& k\`ong             & Free      & Neutral             \\ \hline
	\multirow{2}{*}{\begin{CJK*}{UTF8}{gbsn}假\end{CJK*}} & ji\v a            & Fake      & Neutral/Negative     \\
	& ji\` a            & Holiday   & Neutral             \\ \hline
	\multirow{2}{*}{\begin{CJK*}{UTF8}{gbsn}好吃\end{CJK*}} & h\v aochi             & Delicious & Positive           \\
	& h\`aochi             & Gluttony  & Negative           \\ \hline
\end{tabular}
	}
	
\end{table}

$ $\\Firstly, we come up with two types of phonetic features. The first type extracted audio features from real audio clips. The second type learned pinyin token embeddings from a converted pinyin corpus. For each type of feature, we provide one version with intonation and one version without intonation.

Upon building the feature lookup table between each Chinese pinyin and its feature/embedding, we reach our second step, which is to design a DISA network that works on pinyin sequence and automatically decides the correct intonation for each pinyin. This step is crucial in disambiguating meanings and even sentiment of Chinese characters. Specifically, inspired by~\cite{zhang2018learning}, we employ a reinforcement network as the main structure for our DISA network. The actor network is a typical neural policy network, whose action is to choose one out five intonations for each pinyin. The critic network is an LSTM sequence model, which learns the pinyin sentence sequence representation. The policy network is updated by a delayed reward when the sequence representation is built, while the critic network is updated by a sentiment class cross-entropy loss.
 
 Motivated by the recent success of multimodal learning,  we also incorporate textual and visual features with phonetic features. To the best of our knowledge, we are the first to consider the deep phonemic orthographic characteristic and intonation variation in a multimodal framework for the task of Chinese sentiment analysis. The experimental results show that the proposed multimodal framework outperforms the state-of-the-art Chinese sentiment analysis method by a statistically significant margin. In summary, we make three main contributions in this paper:

\begin{itemize}
	\item We augment the representation of  Chinese characters with additional phonetic cues.
	\item We introduce a reinforcement learning based framework, DISA, which jointly disambiguates intonations of Chinese characters and resolve the sentiment polarity classes of the sentence.
	\item We demonstrate the effectiveness of our framework on several benchmark datasets.
\end{itemize}

The remainder of this paper is organized as follows: we first present a brief review of embedding features, sentiment analysis and Chinese phonetics; we then introduce our model and provides technical details; next, we describe the experimental results and presents analytical discussions; finally, we conclude the paper and suggests future work.

\section{Related Work}\label{sec:related}

We start with a brief review of textual embedding methods. We then analyze existing Chinese representations, which included both textual embeddings and visual-assisted embeddings. Next, we briefly review sentiment analysis and Chinese phonetics.

\subsection{General Embedding}
One-hot representation is the initial numeric word representation method in NLP. However, it usually leads to a problem of high dimensionality and sparsity. To solve this problem, distributed representation (or word embedding)\cite{bengio2003neural} is proposed. Word embedding is a representation which maps words into low dimensional vectors of real numbers by using neural networks. The key idea is based on distributional hypothesis so as to model how to represent context words and the relation between context words and target word. 

In 2013, Mikolov et al.\cite{mikolov2013efficient} introduced both Continuous Bag-of-words model (CBOW) and Skip-gram model. The former placed context words in the input layer and target word in the output layer whereas the latter swapped the input and output in CBOW. In 2014, Pennington et al.~\cite{mikolov2013efficient} created the `GloVe' embeddings. Unlike the previous which learned the embeddings from minimizing the prediction loss, GloVe learned the embeddings with dimension reduction techniques on co-occurrence counts matrix.
\subsection{Chinese Representation}
Chinese text differs from English text for two key aspects: it does not have word segmentations and it has a characteristic of compositionality due to its pictogram nature. Based on the former aspect, word segmentation tools are always employed before text representation, such as ICTCLAS~\cite{zhang2003hhmm}, THULAC~\cite{sun2016thulac}, Jieba\furl{github.com/fxsjy/jieba} and so forth. Based on the latter aspect, several works had focused on the use of sub-word components (such as characters and radicals) to improve word embeddings. \cite{chen2015joint} proposed decomposition of Chinese words into characters and presented a character-enhanced word embedding model (CWE). \cite{sun2014radical,li2015component} had decomposed Chinese characters to radicals and developed a radical-enhanced Chinese character embedding. In \cite{shi2radical}, pure radical based embeddings were trained for short-text categorization, Chinese word segmentation and web search ranking. \cite{yin2016multi} extend the pure radical embedding by introducing multi-granularity Chinese word embeddings.

Multi-modal representation in the past few years has become a growing area of research. \cite{liu2017learning} and \cite{su2017learning} explored integrating visual features to textual word embeddings. The extracted visual features proved to be effective in modeling the compositionality of Chinese characters.

\subsection{Sentiment Analysis and Chinese Phonetics}

Sentiment analysis has raised growing interest both within the scientific community, leading to many exciting open challenges, as well as in the business world, due to the remarkable benefits to be had from financial~\cite{xinfin} and political~\cite{ebrcha} forecasting, user profiling~\cite{mihwha} and community detection~\cite{cavlea}, manufacturing and supply chain applications~\cite{xuuada}, human communication comprehension~\cite{zadatt} and dialogue systems~\cite{youaug}, etc. Various directions have been actively explored in the past few year, from document level~\cite{tang2015document,yang2016hierarchical}, to sentence level~\cite{kim2014convolutional,pang2002thumbs} and to aspect level~\cite{ma2017interactive,peng2018learning}. Most methods took a high perspective to develop effective models for a broad spectrum of languages. Only a limited number of works spend efforts in studying language-specific characteristics~\cite{li2015component,su2017learning,cao2018cw2vec}. Among them, there is almost no literature trying to take advantage of phonetic information for Chinese representation. We, however, believe the Chinese phonetic information could of great value to the representation and sentiment analysis of Chinese language, due to but not limited to the following evidence.

Shu and Anderson conducted a study on Chinese phonetic awareness in~\cite{shu2000phonetic}. The study involved 113 participants of Chinese 2nd, 4th, and 6th graders enrolled in a working-class Beijing, China elementary school. Their task was to represent the pronunciation of 60 semantic phonetic compound characters. Results showed that children as young as 2nd graders are better able to represent the pronunciation of regular characters than irregular characters or characters with bound phonetics. 

The strong influence of familiarity on pronunciation underlines an unavoidable fact about the Chinese writing system: the system does not offer pronunciation cues that are as reliable or consistent as those of many other writing systems, such as English~\cite{albrow1972english}. Moreover, Hsiao and Shillcock argued that semantic-phonetic compound (or phonetic compound) comprised about 81\% of the 7000 frequent Chinese characters~\cite{hsiao2006analysis}. These compounds would affect semantics greatly if we can find an approach to effectively represent their phonetic information.

To this end, no previous work has integrated pronunciation information to Chinese representation. Due to its deep phonemic orthography, we believe the Chinese pronunciation information could elevate the representations to a higher level. Thus, we propose to learn phonetic features and present a DISA network to automatically convert the Chinese character to its pinyin with correct intonation. 

\section{Model}\label{sec:Model}

\label{model}
In this section, we first present how features from textual and visual modalities were extracted. Next, we delve deep into the details of different type of phonetic features. Then, we introduce a DISA network which parses Chinese characters to their pronunciations with tones. Lastly, we demonstrate how we fuse features from three modalities for sentiment analysis.
\subsection{Textual Embedding}
\label{sec:textemb}
As in most recent literature, textual word embedding vectors were treated as the fundamental representation of texts~\cite{mikolov2013efficient,bengio2003neural,pennington2014glove}. Firstly introduced by Bengio et al.~\cite{bengio2003neural}, low-dimensional word embedding vectors learned a distributed representation for words. Compared with traditional n-gram word representations, they largely reduced the data sparsity problem and provided more friendly access towards neural networks. In 2013, Mikolov et al.~\cite{mikolov2013efficient} introduced the toolkit `Word2Vec' which populated the application of word embedding vectors due to its fast learning time. In the toolkit, two predictive-based word vectors, CBOW and Skip-gram, were proposed. They either predicted the target word from context or vice versa. Pennington et al.~\cite{pennington2014glove} developed `GloVe' in 2014 which employed a count-based mechanism to embed word vectors. 
Following the convention, we used `GloVe' character embeddings~\cite{pennington2014glove} of 128-dimension to represent text.

It is worth noting that we set the fundamental token of Chinese text as the character instead of the word for two reasons. Firstly, the character is designed to align against the audio feature. Audio features can only be extracted at character level, as Chinese pronunciation is on each character. In Chinese language, the fundamental phonetic unit which is semantically self-contained is at character level. In English, however, the fundamental phonetic unit is at word level (except some prefix/suffix syllables). Secondly, character-level processing can avoid the errors induced by Chinese word segmentation. Although we used character GloVe embedding as our textual embedding, experimental comparisons were conducted with both CBOW~\cite{mikolov2013efficient} and Skip-gram embeddings.

\subsection{Training Visual Features}
\label{sec:visualfeat}

Unlike the Latin language, the Chinese written language originated from pictograms. Afterwards, simple symbols were combined to form complex symbols in order to express abstract meanings. For example, a geometric combination of three `\begin{CJK*}{UTF8}{gbsn} 木\end{CJK*} (wood)' creates a new character `\begin{CJK*}{UTF8}{gbsn} 森\end{CJK*} (forest)'. 
This phenomenon gives rise to a compositional characteristic of Chinese text. Instead of a direct modeling of text compositionality using sub-word~\cite{chen2015joint,yin2016multi} or sub-character~\cite{li2015component,sun2014radical,penrad} elements, we opt for a visual model. In particular, we constructed a convolutional auto-encoder (convAE) to extract visual features. Details of the convAE are listed in Table~\ref{convae}. 

\begin{table}[]
	\centering
	\caption{Configuration of convAE for visual feature extraction.}
	\label{convae}
	\small
	\begin{tabular}{cc}
		\hline
		Layer\# & Layer Configuration                 \\ \hline
		1       & Convolution 1: kernel 5, stride 1   \\
		2       & Convolution 2: kernel 4, stride 2   \\
		3       & Convolution 3: kernel 5, stride 2   \\
		4       & Convolution 4: kernel 4, stride 2   \\
		5       & Convolution 5: kernel 5, stride 1   \\
		Feature & Extracted visual feature: (1,1,512)  \\
		6       & Dense ReLu: (1,1,1024)  \\
		7       & Dense ReLu: (1,1,2500) \\
		8      & Dense ReLu: (1,1,3600) \\
		9      & Reshape: (60,60,1) \\ \hline
	\end{tabular}
\end{table}

\begin{figure}[]
	\centering
	\includegraphics[width=60mm]{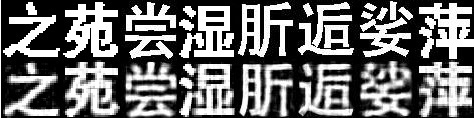} 
	\caption{Original input bitmaps (upper row) and reconstructed output bitmaps (lower row).}
	\label{visualfeat}       
\end{figure}

Following the convention in ~\cite{masci2011stacked} and ~\cite{su2017learning}, we set the input of the model to a 60 by 60 bitmap for each of the Chinese characters and the output of the model to a dense vector with a dimension of 512. The model was trained using Adagrad optimizer on the reconstruction error between original bitmap and reconstructed bitmap. The loss is given as:
\begin{equation}
\sum_{j=1}^{L}(|x_{t}-x_{r}|+(x_{t}-x_{r})^{2})
\end{equation}

where $L$ is the number of samples. $x_{t}$ is the original input bitmap and $x_{r}$ is the reconstructed output bitmap. An example of the original and reconstructed bitmaps is shown in Figure~\ref{visualfeat}. 
After training the visual features, we obtained a lookup table where each Chinese character corresponds to a 512-dimensional feature vector.

\subsection{Learning Phonetic Features}
\label{extractpfeat}
Written Chinese and spoken Chinese have several fundamental differences. To the best of our knowledge, all previous literature that worked on Chinese NLP ignored the significance of the audio channel. As cognitive science suggests, human communication depends not only on visual recognition but also audio activation. This drove us to explore the mutual influence between the audio channel (pronunciation) and textual representation.

Popular Latin and Germanic languages such as Spanish, Portuguese, English etc. share two remarkable characteristics. Firstly, they have shallow phonemic orthography\furl{en.wikipedia.org/wiki/Phonemic_orthography}. In other words, the pronunciation of a word is largely dependent on the text composition in such languages. One can almost infer the pronunciation of a word given its textual spelling. From this perspective, textual information can be interchangeable with phonetic information. 

For instance, if the pronunciations of English word `subject' and `marineland' were known, it is not hard to speculate the pronunciation of word `submarine', because one can combine the pronunciation of `sub' from `subject ' and `marine' from `marineland'. 
This implies that phonetic information of these languages may not have additional information entropy than textual information. Secondly, intonation information is limited and implicit in these languages. Generally speaking, emphasis, ascending intonation and descending intonation are the major variations in these languages. Although they exerted great influence in sentiment polarity during communication, there is no apparent clue to infer such information only from the texts. 

However, Chinese language differs from the above-mentioned languages in several key aspects. Firstly, it is a language of deep phonemic orthography. One can hardly infer the pronunciation of Chinese word/character from its textual writing. For example, the pronunciations of characters `\begin{CJK*}{UTF8}{gbsn}日\end{CJK*}' and `\begin{CJK*}{UTF8}{gbsn} 月\end{CJK*}' are `r\`i' and `yu\`e', respectively. A combination of the two characters makes another character `\begin{CJK*}{UTF8}{gbsn} 明\end{CJK*}' which pronounced `m\'ing'. This characteristic motivates us to find how the pronunciation of Chinese can affect natural language understanding. Secondly, intonation information of Chinese is rich and explicit. In addition to emphasis, each Chinese character has one tone (out of five different tones), marked by diacritics explicitly. These intonations (tones) greatly affect the semantic and sentiment of Chinese characters and words. Examples were shown in Table~\ref{into-tab}.

To this end, we found it was not trivial to explore how Chinese pronunciation can influence natural language understanding, especially sentiment analysis. In particular, we designed two approaches to learn phonetic information, namely feature extraction from audio signal and embedding vector learning from textual corpus. For either of the above two approaches, we have two variations, namely with (Ex04, PW) or without (Ex0, PO) intonations. An illustration is shown in Table~\ref{pinyinstat1}. Details of each type will be introduced in the following sections.
\begin{table}[]
	\centering
	\caption{Illustration of 4 types of phonetic features: a(x) stands for the extracted audio feature for pinyin `x'; v(x) represents learned embedding vector for `x'; number 0 to 4 represents 5 diacritics.}
	\label{pinyinstat1}
	\resizebox{0.5\textwidth}{!}{%
		\begin{tabular}{c|l|l}
			\hline
			\multicolumn{2}{c|}{Text}                                                              & \begin{CJK*}{UTF8}{gbsn}假设明天放假。\end{CJK*}                                         \\
			\multicolumn{2}{c|}{English}                                                           & Suppose tomorrow is holiday.                 \\ \hline
			\multicolumn{2}{c|}{Pinyin}                                                            & Ji\v a Sh\`e M\'ing Ti\=an F\`ang Ji\`a                                         \\ \hline
			\multirow{2}{*}{\begin{tabular}[c]{@{}c@{}}Extracted\\ from audio\end{tabular}} & Ex0  & a(Jia) a(She) a(Ming) a(Tian) a(Fang) a(Jia) \\
			& Ex04 & a(Ji\v a) a(Sh\`e) a(M\'ing) a(Ti\=an) a(F\`ang) a(Ji\`a)                                         \\ \cline{1-2}
			\multirow{2}{*}{\begin{tabular}[c]{@{}c@{}}Learned\\ from corpus\end{tabular}}  & PO   & v(Jia) v(She) v(Ming) v(Tian) v(Fang) v(Jia)                   \\
			& PW   & v(Jia3) v(She4) v(Ming2) v(Tian1) v(Fang4) v(Jia4)             \\ \hline
		\end{tabular}
	}
\end{table}

\subsubsection{Extracted feature from audio clips (\textbf{Ex0, Ex04})}
\label{sec:ex}
The spoken system of modern Chinese is named `Hanyu Pinyin', abbreviated to `pinyin'\furl{iso.org/standard/13682.html}. It is the official romanization system for mandarin in mainland China~\cite{benjamin1997history}. The system includes four diacritics denoting four different tones plus one neutral tone. For each of the Chinese characters, it has one corresponding pinyin. This pinyin has five variations in tones (we treat the neutral tone as one special tone). The statistics of Chinese character and pinyin are listed in Table~\ref{pinyinstat2}.
It shows that the number of frequently used characters is bigger than the number of pinyins with or without tones. 
This suggests that certain Chinese characters share the same pinyin and further implies that the one-hot dimensionality will reduce if pinyin was used to represent text.

In order to extract phonetic features, for each tone of each pinyin, we collected an audio clip which recorded a female's pronunciation of that pinyin (with tone) from a language learning resource\footnote{http://chinese.yabla.com -- This resource has only four tones for each pinyin, which does not have the neutral tone pronunciation. To obtain the neutral tone feature, we compute the arithmetic mean of the features of the other four tones.}. Each audio clip lasts around one second with a standard pronunciation of one pinyin with tone. The quality of these clips was validated by two native speakers. Next, we used	openSMILE~\cite{eyben2010opensmile} to extract phonetic features on each of the obtained pinyin-tone audio clip. Audio features are extracted at 30 Hz frame-rate and a sliding window of 20 ms. They consist of a total number of 39 low-level descriptors (LLD) and their statistics, e.g., MFCC, root quadratic mean, etc.

\begin{table}[]
	\centering
	\caption{Statistics of Chinese characters and `Hanyu Pinyin'}
	\label{pinyinstat2}
	\resizebox{0.5\textwidth}{!}{%
\begin{tabular}{c|cc|cl}
	\hline
	\multirow{2}{*}{} & \multicolumn{2}{c|}{Pinyin} & \multicolumn{2}{c}{\multirow{2}{*}{\begin{tabular}[c]{@{}c@{}}Textual\\ Character\end{tabular}}} \\ \cline{2-3}
	& w/o tones     & w/ tones    & \multicolumn{2}{c}{}                                                                             \\ \hline
	Number of tokens  & 374           & 1870        & \multicolumn{2}{c}{3500}                                                                         \\ \hline
\end{tabular}
	}
\end{table}

After obtaining features for each of the pinyin-tone clip, we obtained an $m\times39$ dimensional matrix for each clip, where $m$ depends on the length of clip and $39$ is the number of features. To regulate the feature representation for each clip, we conducted Singular value decomposition (SVD) on the matrices to reduce them to 39-dimensional vectors, where we extracted the vector with the singular values. In the end, high dimensional feature matrices of each pinyin clip were transformed to a dense feature vector of 39 dimensions. A lookup table between pinyin and audio feature vector is constructed accordingly.

In particular, we prepared two sets of extracted phonetic features. The first type comes with tone, which is the feature we obtained from the above processing. We denote it as \textbf{Ex04}, where `Ex' stands for extracted features and `04' stands for having one tone from 0 to 4 (we represent neutral tone as 0 and the first to the fourth tone as 1 to 4 respectively). The second type removed the variations of tones, in which we take the arithmetic mean of five features from five tones of each pinyin. We denote it as \textbf{Ex0}, where `0' stands for no tone. In the second type of feature, pinyins with different tones will have same phonetic features, even though they may mean different meanings. 

\begin{figure*}[]
	\centering
	\small
	\resizebox{1\textwidth}{!}{%
		\includegraphics[]{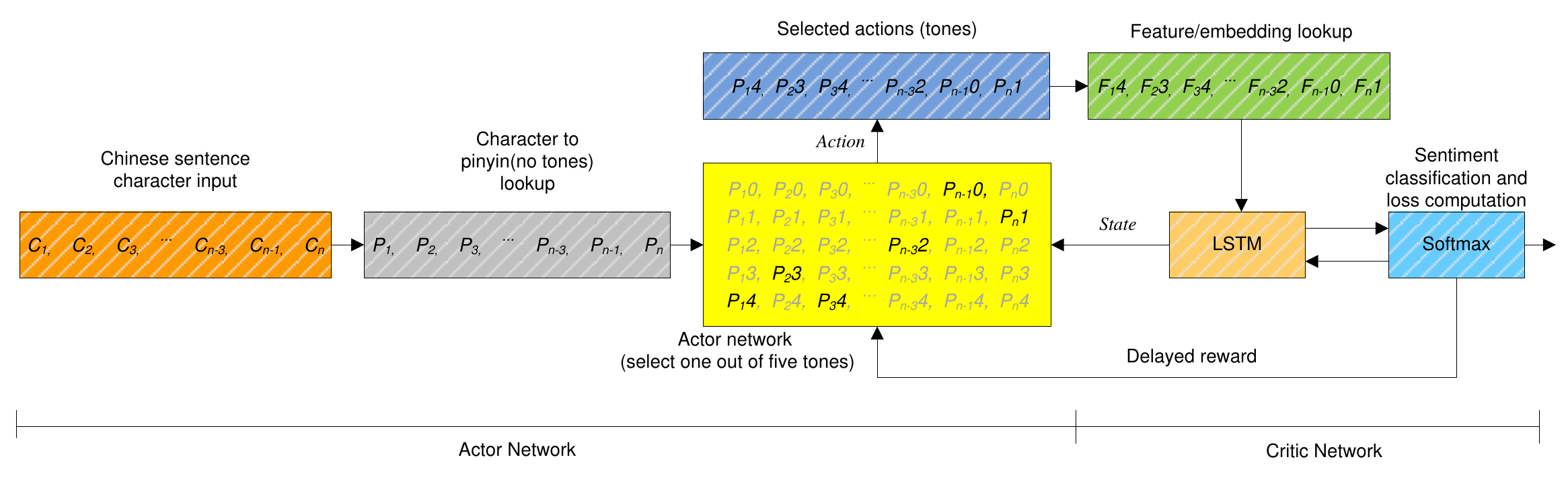} 
	}
	\caption{DISA model structure for tone selection. $C_{m}$ stands for the $m$th Chinese character in a sentence. $P_{m}$ denotes the pinyin for $m$th character without the tones. $P_{m}n$ represents the pinyin for $m$th character with its $n$th tone. $F_{m}n$ is the feature/embedding vector for $P_{m}n$.}
	\label{frame}       
\end{figure*}

\subsubsection{Learned feature from pinyin corpus (\textbf{PO, PW})}
\label{sec:popw}
Instead of collecting audio clips for each pinyin and extracting audio features, we directly represent Chinese characters with pinyin tokens, as shown in Table~\ref{pinyinstat1}. Specifically, we convert each Chinese character in a textual corpus to it pinyin. The original corpus which was represented by a sequence of Chinese characters was converted to a phonetic corpus which was represented by a sequence of pinyin tokens. 

In the phonetic corpus, contextual semantics were still maintained as in textual corpus. This is achieved with the help of online parser\furl{github.com/mozillazg/python-pinyin}, which parse Chinese characters to their pinyins. It should be pointed out that 3.49\% of the common 3500 Chinese characters (around 122 characters) have multiple pinyins\furl{yyxx.sdu.edu.cn/chinese/new/content/1/04/main04-03.htm}, namely `duo yin zi'(heteronym). Although the parser claimed its support to heteronym, we took the most statistically-possible pinyin prediction of each heteronym.

 We did not disambiguate various heteronyms particularly, as this is not the major assumption we try to argue in this paper. However, it could be a direction worth working on in the future. The DISA provides two modes in its conversion from Character to pinyin, one with tone and the other without tone. 

For the mode without tone, Chinese characters will be converted to pinyin without tones only. Examples are the tokens shown in the row of PO in Table~\ref{pinyinstat1}, where \textbf{PO} stands for Pinyin w/o tones. Afterwards, we train 128-dimension pinyin token embedding vectors using conventional `GloVe' character embeddings~\cite{pennington2014glove}. A lookup table between pinyin without intonation (\textbf{PO}) and embedding vector is constructed accordingly. Pinyins that have the same pronunciation but different intonations will share the same glove embedding vector, such as \textit{Ji\v a} and \textit{Ji\`a} in Table~\ref{pinyinstat1}.

For the mode with tone, Chinese characters will be converted to pinyin plus a number suggesting the tone. Examples are the tokens shown in the row of PW in Table~\ref{pinyinstat1}, where \textbf{PW} stands for Pinyin w/ tones. We use number 1 to 4 to represent four diacritics and number 0 to represent the neutral tone. Similarly, 128-dimension `GloVe' pinyin embedding vectors were trained.

In sum, we have four types of phonetic features, namely \textbf{Ex04}, \textbf{PW}, \textbf{E0} and \textbf{PO}. PO distinguishes from PW in removing intonations. Two of them (\textbf{Ex04}, \textbf{PW}) distinguish from others by having intonations. It is expected to have one question that how would one know the correct intonation of pinyins given their textual characters. Although the online parser can give its statistical guess, the performance and robustness can not be evaluated and guaranteed. To address this problem, we design a parser network with a reinforcement learning model to learn the correct intonation of each pinyin. Details will be presented in the following section.

\subsection{DISA}
\subsubsection{Overview}
This DISA network takes a sentence of Chinese characters as input. It firstly converts each character to its corresponding pinyin (without tones) through a lookup operation. Then the pinyin sequence will be fed to an \textbf{actor-critic network}. For each pinyin (time step), a policy network will randomly sample one out of five actions, where each action denotes a tone. Then a feature/embedding of this specific pinyin with tone is retrieved from a feature lookup module. 

During exploration stage, the action will be randomly sampled. During exploitation and prediction stages, the action will be the one with maximum probability given the policy. This feature/embedding sequence will then be fed to an LSTM network. Hidden states from the LSTM will pass back to policy network for guiding action selection. The final hidden state of the LSTM network will be fed to a softmax classifier to obtain a sentence sentiment class distribution. A log probability of ground-truth label will be treated as a delayed reward to tune the policy network. Finally, a cross entropy loss will be computed against the obtained sentiment class distribution to tune the critic network. A graphical description is shown in Figure~\ref{frame}, followed by details below.

\textbf{State}: For the environment, we used an LSTM to simulate the value function (detailed later). The input to this LSTM is the sequence of feature/embedding retrieved from the lookup module (detailed later), namely $x_1, x_2,...x_t, ..., x_L$, where $x_t$ is the feature for the t$th$ pinyin in the sentence. The mathematical representations of the LSTM cell are as follows:

\begin{equation}
\label{eq:lstm}
\begin{aligned}
\small
f_{t} &=\sigma (W_{f} [x_t,h_{t-1}] + b_{f}) \\
I_{t} &=\sigma (W_{I} [x_t, h_{t-1}] + b_{I})\\
\widetilde{C}_{t} &=tanh (W_{C} [x_t,h_{t-1}] + b_{C})\\
C_{t} &= f_{t}* C_{t-1} +I_{t}* \widetilde{C}_{t} \\
o_{t} &=\sigma (W_{o} [x_t,h_{t-1}] + b_{o}) \\
h_{t} &= o_{t}* tanh({C}_{t}) \\
\end{aligned}
\end{equation}
where $f_{t}$, $I_{t}$ and $o_{t}$ are the forget gate, input gate and output gate, respectively. $W_{f}$, $W_{I}$,$W_{o}$, $b_{f}$, $b_{I}$ and $b_{o}$ are the weight matrix and bias scalar for each gate. ${C}_{t}$ is the cell state and $h_{t}$ is the hidden state output. 

The state of the environment is defined as:

\begin{equation}
\label{for_state}
S_{t}=[x_{t}\oplus h_{t-1}\oplus C_{t-1}]
\end{equation}
where $\oplus$ is a concatenation (same below). As shown in Formula~\ref{for_state}, the state is determined by the current feature input, the last LSTM hiddent output and the last LSTM cell memory.

\textbf{Action}:
There are five actions in our environment, representing five different tones. An example is shown in Table~\ref{act_tone}. If different action was selected, then the corresponding intonation will be activated. Relevant phonetic features will then be selected, as introduced in Section~\ref{seclookup}. The action policy was implemented by a typical feedforward neural network. Specifically, for a policy $\pi ({a_{t}\mid S_{t}})$ at time $t$,

\begin{equation}
\label{for_policy}
\pi ({a_{t}\mid S_{t}})=\tanh (W \cdot S_{t} + b)
\end{equation}

where $W$ and $b$ are the weight matrix and bias scalar. $a_{t}$ is the action at time $t$. During exploration of training, action will be randomly selected out of the above five. During exploitation of training and testing, the action with the maximum probability will be selected.

\begin{table}[]
\centering
	\caption{Actions in DISA network and meanings.}
		\label{act_tone}
		\resizebox{0.45\textwidth}{!}{%
	\begin{tabular}{c|ccccc}
		\hline
		Action     & 0    & 1 & 2 & 3 & 4 \\ \hline
		Intonation & Neutral & \=.    & \' .   & \v.    & \`.    \\ \hline
		Example    & a    & \= a & \' a & \v a & \` a  \\ \hline
	\end{tabular}
}
\end{table}

\textbf{Reward}:
The reward is computed by the end of each sentence when the state/action trajectory comes to the terminal (delayed reward). After the feature/embedding lookup module, the feature sequence is fed to the LSTM critic network. A sentence sentiment class distribution is computed as:

\begin{equation}
\label{for_softmax}
distr=\sigma (W_{sfmx} \cdot h_{L} + b_{sfmx})
\end{equation}

where $W_{sfmx}$ and $b_{sfmx}$ are weight matrix and bias scalar from the softmax layer. $h_{L}$ is the last hidden state output from the LSTM critic network. $distr^{1*X}$ is the probability distribution of sentiment classes for the sentence. $X$ is the number of sentiment class. The reward ($R$) is defined as:

\begin{equation}
\label{for_reward}
R=log(P(ground \mid sent))
\end{equation}

where $P(ground \mid sent)$ stands for the probability of the ground-truth label of the sentence given the distribution in Eq.~\ref{for_softmax}.

\subsubsection{Actor: policy network}
As shown in the `Action' above, the policy network random guesses actions during the exploration stage in training. It will be updated when a sentence input is fully traversed. Given the reward obtained from Eq.~\ref{for_reward}, we used gradient descent method to optimize the policy network~\cite{sutton2000policy}. In other words, we want to maximize:

\begin{equation}
\label{for_obj}
\begin{aligned}
\small
J(\theta ) &=E_{\pi }[R(S_{1},a_{1},S_{2},a_{2},...,S_{L},a_{L})] \\
 &= \sum_{1}^{L}p(S_{1})\prod_{t}\pi _{\theta }({a_{t}\mid S_{t}})p(S_{t+1}\mid S_{t},a_{t}) R_{L} \\
 &= \sum_{1}^{L}\prod_{t}\pi _{\theta }({a_{t}\mid S_{t}}) R_{L}\\
\end{aligned}
\end{equation}

Using the likelihood ratio (or REINFORCE~\cite{williams1992simple} trick) to estimate policy gradient, the gradient can be transformed to:

\begin{equation}
\label{for_gra}
\nabla _{\theta }J(\theta )=\sum_{t=1}^{L}R_{L}\nabla _{\theta }log \pi _{\theta }(a_{t}\mid S_{t})
\end{equation}

\begin{figure}[]
	\centering
	\resizebox{0.5\textwidth}{!}{%
		\includegraphics[]{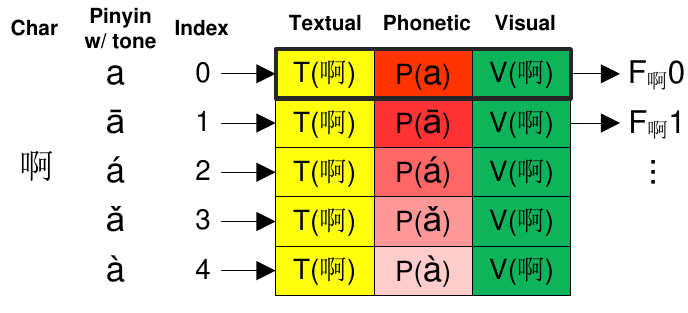} 
	}
	\caption{An example of fused character feature/embedding lookup, where T, P, V represent features/embeddings from corresponding modality. In the case of single modality or bi-modality, relevant lookup table is constructed accordingly.}
	\label{fusionlookup}       
\end{figure}

\subsubsection{Feature/embedding lookup} \label{seclookup}

Recall that we have selected actions from actor network, where each action denotes a tone for that pinyin, the function of this feature/embedding lookup module is to retrieve the correct features of that specific pinyin with tone. Prior to the policy network, we have collected phonetic features from five different tones of each pinyin and order them from neutral tone feature to the fourth tone feature. The neutral tone to the fourth tone feature can be retrieved individually by index ID number 0 to 4.

When an action is selected from the actor network, for example, action 4 was selected for pinyin $P_{1}$, this lookup module will find the fourth phonetic feature (index ID 4) of this pinyin, namely $F_{1}4$ and pass it to the LSTM critic network as the input $x_t$ in Eq.~\ref{eq:lstm}.

\subsubsection{Critic: sentence model and loss computation}

Introduced in the \textbf{State} before, the critic network was essentially a sentence encoding model by an LSTM. We used gradient descent method to update the critic network with the cross-entropy loss defined as:

\begin{equation}
\label{for_entropy}
L=-\sum_{\forall sent}P(ground \mid sent)log(P(pred \mid sent))
\end{equation}

where $P(ground \mid sent)$ and $P(pred \mid sent)$ are the ground truth and predicted probability in the Eq.~\ref{for_softmax}, respectively.

\subsection{Fusion of Modalities}
In the context of the Chinese language, textual embeddings have been applied in various tasks and proved its effectiveness in encoding semantics or sentiment~\cite{chen2015joint,li2015component,sun2014radical,shi2radical,yin2016multi}. Recently, visual features pushed the performance of textual embedding further via a multimodal fusion~\cite{su2017learning,liu2017learning}. This is achieved due to the effective modeling of compositionality of Chinese characters by the visual features. In this work, we hypothesize that the use of phonetic features along with textual and visual can improve the performance. Thus, we introduced the following fusion method that fits with our DISA network, as in Figure~\ref{frame}.

\begin{itemize}
	\item Each Chinese character is represented by a concatenation of three segments. Each segment represents one modality, see below:
	
		\begin{equation}
		char=[emb_{T}\oplus emb_{P}\oplus emb_{V}]
		\end{equation}
		where $char$ is character representation. $emb_{T}, emb_{P}, emb_{V}$ are embeddings from text, phoneme and vision, respectively.
\end{itemize}

There are other complex fusion methods available in the literature~\cite{zadeh2017tensor}, however, we did not use them in our paper for three reasons. (1) Fusion through concatenation is one proven effective method~\cite{snoek2005early,karpathy2014large,liu2017learning}, and (2) it has the added benefit of simplicity, thus allowing for the emphasis (contributions) of the system to remain with the features themselves. (3) The designed fusion needs to fit in with our reinforcement model framework. Fusion methods as in~\cite{su2017learning} and ~\cite{zadeh2017tensor} impose obstacles in the implementation with actor-critic model. Thus, we used the above introduced fusion method, an example of a fused feature/embedding lookup table is shown in Fig.~\ref{fusionlookup}.

\section{Experiments and Results}\label{sec:experiments}

In this section, we start with introducing the experimental setup. Experiments were conducted in six steps. Firstly, we compare unimodal features. Secondly, we experiment on the possible fusion of modalities. Thirdly, we compare cross domain validation performance between our method with baselines. Next, we conduct ablation tests to validate the contribution of phonetic features. More precisely, we also visualize different phonetic features/embeddings to understand how they improve the performance.

\subsection{Experimental Setup}

\subsubsection{Datasets and features/embeddings}

\textbf{Datasets}: We evaluate our method on five datasets: Weibo, It168, Chn2000, Review-4 and Review-5. The first three datasets consist of reviews extracted from micro-blog and review websites. The last two datasets contain reviews from~\cite{che2015sentence}, where Review-4 has reviews from computer and camera domains, and Review-5 contains reviews from car and cellphone domains. The experimental datasets are shown in Table~\ref{meta}. 

\textbf{Features/embeddings}: For textual embeddings, we refer to the pretrained character embedding lookup table trained with Glove in Section~\ref{sec:textemb}. For phonetic experiments, we employ a pre-built tool called online codes\furl{github.com/mozillazg/python-pinyin} on the datasets to convert text to pinyin without intonations (As we discussed in Section~\ref{sec:popw}, this conversion achieves as high as 97\% accuracy.). Ex0 and Ex04 features were extracted from audio files and stored as in Section~\ref{sec:ex}. PO and PW embeddings were also pretrained on the same textual corpus for training textual embeddings. The corpus contains news of 8
million Chinese words, which is equal to 38 million Chinese characters. For visual features, we refer to the lookup table to convert characters to visual features as in Section~\ref{sec:visualfeat}.

For experiments of multimodality, features from each individual modality were concatenated into a lookup table. Examples are shown in Fig.~\ref{fusionlookup}.

\begin{table}[h]
	\centering
	\caption{Statistics of experimental datasets}
	\label{meta}
	\resizebox{0.5\textwidth}{!}{%
		\begin{tabular}{cccccc}
			\hline
			& Weibo & It168 & Chn2000 & Review-4 & Review-5 \\ \hline
			Positive & 1900  & 560    & 600   & 1975      & 2599      \\
			Negative & 1900  & 458    &  739  & 879       & 1129      \\
			Sum      & 3800  & 1018    &  1339 & 2854      & 3728      \\ \hline
		\end{tabular}
	}
\end{table}


\subsubsection{Setup and Baselines}

\textbf{Setup}: We use TensorFlow and Keras to implement our model. All models use an Adam Optimizer with a learning rate of 0.001 and an L2-norm regularizer of 0.01. Dropout rate is 0.5. Each mini-batch contains 50 samples. We split each dataset to training, testing and development sets per the ratio 6:2:2. We report the result of testing set whose corresponding development set performs the best after 30 epochs. The above parameters were set with the use of a grid search on the development data.

The training procedure of our DISA network is as follows. Firstly, we skip the policy network and directly train the LSTM critic network with the training objective as Eq.~\ref{for_entropy}. Secondly, we fix the parameters of the LSTM critic network and train the policy network with the training objective as Eq.~\ref{for_gra}. Lastly, we co-train all the modules together until convergence. For the cases when no phonetic feature/embedding is involved, for example pure textual or visual features, only the LSTM is trained and tested. Glove was chosen as the textual embedding in our model due to its performance in Table~\ref{s-mod-tab}.

\textbf{DISA variants}: We introduce below the variants of our DISA network. They differ in text representation features.

\begin{enumerate}

	\item \textbf{DISA (P)}: DISA network that used phonetic feature only, which is the concatenation of Ex04 and PW.
	\item \textbf{DISA (T+P)}: DISA network that uses the concatenation of textual embedding (glove) and phonetic feature (Ex04+PW).
	\item \textbf{DISA (P+V)}: DISA network that uses the concatenation of phonetic feature (Ex04+PW) and visual feature.
	\item \textbf{DISA (T+P+V)}: DISA network that uses the concatenation of textual embedding (glove), phonetic feature (Ex04+PW) and visual feature.

\end{enumerate}

\textbf{Baselines}: 
Our proposed method is based on input features/embeddings of Chinese characters. 
In related works of Chinese textual embedding, all of them aimed at improving Chinese word embeddings, such as CWE~\cite{chen2015joint}, MGE~\cite{yin2016multi}. Those who utilized visual features~\cite{su2017learning,liu2017learning} also aimed at word level. However, they cannot stand as fair baselines to our proposed model, as our model studies Chinese character embedding. There are two major reasons for studying at character level. Firstly, pinyin pronunciation system is designed for character level. Pinyin system does not have corresponding pronunciations to Chinese words. Secondly, character level will bypass Chinese word segmentation operation which may induce errors. Conversely, using character level pronunciation to model word level pronunciation will incur sequence modeling issues. For instance, a Chinese word `\begin{CJK*}{UTF8}{gbsn}你好\end{CJK*}' is comprised of two characters, `\begin{CJK*}{UTF8}{gbsn}你\end{CJK*}' and `\begin{CJK*}{UTF8}{gbsn}好\end{CJK*}'. For textual embedding, the word can be treated as one single unit by training a word embedding vector. For phonetic embedding, however, we cannot treat the word as one single unit from the perspective of pronunciation. The correct  pronunciation of the word is a time sequence of character pronunciation of firstly `\begin{CJK*}{UTF8}{gbsn}你\end{CJK*}' and then `\begin{CJK*}{UTF8}{gbsn}好\end{CJK*}'. If we work at word level, we have to come up with a representation of the pronunciation of this word, such as an average of character phonetic features etc.
To make a fair comparison, we compare with character level methods below:

\begin{enumerate}
	\item \textbf{Glove}: An unsupervised embedding learning algorithm based on co-occurrence (count).~\cite{pennington2014glove}. 
	
	\item  \textbf{CBOW}: Continuous Bag-of-words model which places context words in the input layer and target word in the output layer~\cite{mikolov2013efficient}.
	\item  \textbf{Skip-gram}: The opposite of CBOW model, which predicts the contexts given the target word~\cite{mikolov2013efficient}.
	\item  \textbf{Visual}: Based on~\cite{su2017learning} and~\cite{masci2011stacked}, a convolutional auto-encoder (convAE) is built to extract compositionality of Chinese characters through the visual channel.
	\item  \textbf{charCBOW}: Component-enhanced character embedding built on top of CBOW method by~\cite{li2015component}. It delved into the radical components of Chinese characters and enriched the character representation with radical component.
	\item  \textbf{charSkipGram}: The Skip-gram varaint of charCBOW.
	\item  \textbf{Hsentic}: It proposed radical-based hierarchical embeddings for Chinese sentiment analysis. Character representations were specifically tuned for sentiment analysis~\cite{penrad}.
	
\end{enumerate}


\begin{table}[t]
	\centering
	\small
	\caption{Classification accuracy of unimodality in LSTM.}
	\label{s-mod-tab}
	\resizebox{0.5\textwidth}{!}{%
		
\begin{tabular}{ccccccc}
	\hline
	\multicolumn{2}{c}{}                                              & Weibo          & It168          & Chn2000        & Review-4       & Review-5       \\ \hline
	\multicolumn{2}{c}{GloVe}                                         & \textbf{75.39} & 81.82          & \textbf{84.54} & \textbf{87.46} & \textbf{86.94} \\
	\multicolumn{2}{c}{CBOW}                                          & 72.39          & 78.75          & 81.18          & 85.11          & 84.71          \\
	\multicolumn{2}{c}{Skip-gram}                                     & 75.05          & 80.13          & 78.04          & 86.23          & 86.21          \\
	\multicolumn{2}{c}{Visual}                                        & 61.78          & 65.40          & 67.21          & 78.98          & 79.59          \\
	\multicolumn{2}{c}{charCBOW}                                      & 71.54          & 80.83          & 82.82          & 86.90          & 85.19          \\
	\multicolumn{2}{c}{charSkipGram}                                  & 71.86          & 82.10          & 81.63          & 85.21          & 84.84          \\
	\multicolumn{2}{c}{Hsentic}                                       & 73.65          & 80.23          & 79.09          & 84.76          & 73.31          \\ \hline
	\multicolumn{1}{c|}{\multirow{3}{*}{Phonetic features}} & DISA(Ex04)    & 67.28          & 84.69          & 78.18          & 81.88          & 83.38          \\
	\multicolumn{1}{c|}{}                                   & DISA(PW)      & 67.80          & 83.73          & 77.45          & 85.37          & 84.18          \\
	\multicolumn{1}{c|}{}                                   & DISA(P) & 68.19          & \textbf{85.17} & 79.27          & 84.67          & 85.24          \\ \hline
\end{tabular}
	}
\end{table}

\subsection{Experiments on Unimodality}

For textual embeddings, we have compared with state-of-the-art embedding methdos including GloVe, skip-gram, CBOW, charCBOW, charSkipGram and Hsentic. As shown in Table~\ref{s-mod-tab}, textual embeddings (GloVe) achieve the best performance among all three modalities in four datasets. This is due to the fact that they successfully encoded the semantics and dependency between characters. We also find that charCBOW and charSkipGram methods perform quite close to the original CBOW and Skip-gram methods. They perform slightly but not constantly better than their baselines. We conjecture this could be caused by the relatively small size of our training corpus compared to the original Chinese Wikipedia Dump training corpus. With the corpus size increased, all embedding methods are expected to have improved performance. It is without doubts, though, that the corpus we used still presents a fair platform for all methods to compare. 

We also notice that visual feature achieve the worst performance among three modalities, which is within our expectation. As demonstrated in ~\cite{su2017learning}, pure visual features are not representative enough to obtain a comparable performance with the textual embedding. Last but not least, our methods with phonetic features perform better than visual feature. Although visual features capture compositional information of Chinese characters, they fail to distinguish different meanings of characters that have same writing but different tones. These tones could largely alter the sentiment of Chinese words and further affect sentiment of sentence.

For phonetic representation, three types of features were tested, namely EX04, PW and P (namely EX04+PW). The last one is the concatenation of the previous two. Our first observation is that phonetic features alone can hardly compete with textual embeddings. Although they beat textual embeddings in It168 dataset, they consistently fell behind textual embeddings. This is still within our expectation, as suggested by Tseng in~\cite{tseng1983acoustic}, `Phonology and phonetics alone are insufficient in predicting the actual output of sentences'.

If we further refer to the Table~\ref{pinyinstat2}, we can find that on average 2 to 3 characters share one same pinyin with tone. That means a pure phonetic representation may wipe out the 1 out of 2 or 3 (33\%-50\%) semantics from the text. This inevitably will reduce the possibility to correctly classify the sentiment.

As we can see each modality has its own capacity to encode semantics, it is expected to take advantage of the complimentary information from multiple modalities for the sentiment analysis task. The results are shown in the next section. 


\subsection{Experiments on Fusion of Modalities}
In this set of experiments, we evaluate the fusion of every possible combination of modalities. After extensive experimental trials, we summarize that the concatenation of Ex04 and PW embeddings (denoted as P) performed best among all phonetic feature combinations. Thus we use it as phonetic feature in the fusion of modalities. The results shown in Table~\ref{m-mod-tab} suggest that the best performance is achieved by fusing textual and phonetic features. 

\begin{table}[t]
	\centering
	\caption{Classification accuracy of multimodality. (T and V represent textual and visual, respectively. + means the fusion operation. P is the concatenated phonetic feature of the one extracted from audio (Ex04) and Pinyin w/ intonation (PW).)}
	\label{m-mod-tab}
	\small
	\resizebox{0.5\textwidth}{!}{%
\begin{tabular}{clccccc}
	\hline
	\multicolumn{2}{c}{}              & Weibo          & It168          & Chn2000        & Review-4       & Review-5       \\ \hline
	\multicolumn{2}{c}{GloVe}         & 75.39          & 81.82          & 84.54          & 87.46          & 86.94          \\
	\multicolumn{2}{c}{Visual}        & 61.78          & 65.40          & 67.21          & 78.98          & 79.59          \\
	\multicolumn{2}{c}{charCBOW}      & 71.54          & 80.83          & 82.82          & 86.90          & 85.19          \\
	\multicolumn{2}{c}{charSkipGram}  & 71.86          & 82.10          & 81.63          & 85.21          & 84.84          \\
	\multicolumn{2}{c}{Hsentic}       & 73.65          & 80.23          & 79.09          & 84.76          & 73.31          \\
	\multicolumn{2}{c}{DISA(P)}       & 68.19          & 85.17          & 79.27          & 84.67          & 85.24          \\ \hline
	\multicolumn{2}{c}{DISA(T+P)}    & \textbf{75.75} & \textbf{86.12} & \textbf{85.45} & \textbf{90.42} & \textbf{90.03} \\
	\multicolumn{2}{c}{DISA(T+V)}          & 73.79          & 85.65          & 83.27          & 89.37          & 88.70          \\
	\multicolumn{2}{c}{DISA(P+V)}    & 76.01          & 82.30          & 81.09          & 86.76          & 87.23          \\
	\multicolumn{2}{c}{DISA(T+P+V)} & 74.32          & 77.99          & 78.18          & 87.63          & 89.49          \\ \hline
\end{tabular}
	}
	\vspace{-0.5cm}
\end{table}

We notice that phonetic features when fused with textual or visual features, improve the performance of both textual and visual unimodal classifiers consistently. This validates our hypothesis that phonetic features are an important factor in representing semantics, which leads to an improvement in Chinese sentiment analysis performance. A p-value of 0.007 in the paired t-test between with and without phonetic features suggested that the best performing improvement of integrating phonetic feature is statistically significant. The integration of multiple modalities could take advantages of information from different modalities. However, we notice that, in most of the cases, tri-modal models underperform bi-modal models. One disadvantage brought by using more modalities is the increase of number of parameters. We conjecture that a larger set of learnable parameters leads to poor generalizability when the training sets in our experiments only consist of instances of less than 4000. 

Furthermore, the information redundancy becomes more severe when combining features across different modalities. In other words, there might be the marginal effect of using additional modality. We will illustrate this point with examples. As aforementioned, Chinese character is made of symbols (or called radicals). Some symbols function as morphemes, while some symbols function as phonemes. For instance, the character \begin{CJK*}{UTF8}{gbsn}`疯'\end{CJK*} consists of two symbols, \begin{CJK*}{UTF8}{gbsn}`疒'\end{CJK*} and \begin{CJK*}{UTF8}{gbsn}`风'\end{CJK*}. The pronunciation of \begin{CJK*}{UTF8}{gbsn}`疯'\end{CJK*} (feng1) is dominated by the symbol \begin{CJK*}{UTF8}{gbsn}`风'\end{CJK*} (feng1), which is the same for phonetic features. Meanwhile, \begin{CJK*}{UTF8}{gbsn}`风'\end{CJK*} contributes the most to the visual image of \begin{CJK*}{UTF8}{gbsn}`疯'\end{CJK*}, the visual feature of \begin{CJK*}{UTF8}{gbsn}`疯'\end{CJK*} can also somehow encodes the information brought by \begin{CJK*}{UTF8}{gbsn}`风'\end{CJK*}. 

After we compare T with T+P and T+V, the performance increase induced by P is 1.40\% higher than by V on average. It is apparent to conclude that phonetic feature is better at encoding semantics than visual features. The fusion of phonetic and textual embeddings achieve the best performance in all of the cases. It indicates that the information encoded in the phonetic feature complements that of textual embedding.

\begin{table}[t]
	\centering
	\caption{Cross-domain evaluation. Datasets on the first column are the training sets. Datasets on the first row are the testing sets. The second column represents various baselines and our proposed method.}
	\label{cross-tab}
	\resizebox{0.5\textwidth}{!}{%
		
		\begin{tabular}{cc|ccccc}
			\hline
			&              & Weibo              & It168              & Chn2000            & Review-4           & Review-5           \\ \hline
			\multicolumn{1}{c|}{\multirow{4}{*}{Weibo}}    & Hsentic      & \multirow{4}{*}{-} & 66.47              & 61.84              & 64.93              & 63.71              \\
			\multicolumn{1}{c|}{}                          & charCBOW     &                    & 67.55              & 64.08              & 62.09              & 67.78              \\
			\multicolumn{1}{c|}{}                          & charSkipGram &                    & 65.29              & 59.60              & 53.22              & 49.49              \\
			\multicolumn{1}{c|}{}                          & DISA(T+P)    &                    & \textbf{73.68}     & \textbf{66.55}     & \textbf{69.16}     & \textbf{71.01}     \\ \hline
			\multicolumn{1}{c|}{\multirow{4}{*}{It168}}    & Hsentic      & 59.15              & \multirow{4}{*}{-} & 59.30              & 69.76              & 67.62              \\
			\multicolumn{1}{c|}{}                          & charCBOW     & 57.54              &                    & 65.05              & 72.25              & 68.13              \\
			\multicolumn{1}{c|}{}                          & charSkipGram & 54.54              &                    & 64.68              & 68.19              & 64.38              \\
			\multicolumn{1}{c|}{}                          & DISA(T+P)    & \textbf{63.75}     &                    & \textbf{68.36}     & \textbf{77.00}     & \textbf{74.07}     \\ \hline
			\multicolumn{1}{c|}{\multirow{4}{*}{Chn2000}}  & Hsentic      & 56.36              & 60.67              & \multirow{4}{*}{-} & 52.03              & 44.77              \\
			\multicolumn{1}{c|}{}                          & charCBOW     & 56.23              & 70.40              &                    & 61.77              & 63.36              \\
			\multicolumn{1}{c|}{}                          & charSkipGram & 51.99              & 68.53              &                    & 62.47              & 62.77              \\
			\multicolumn{1}{c|}{}                          & DISA(T+P)   & \textbf{60.50}     & \textbf{68.90}     &                    & \textbf{68.64}     & \textbf{69.02}     \\ \hline
			\multicolumn{1}{c|}{\multirow{4}{*}{Review-4}} & Hsentic      & 58.15              & 73.55              & 59.22              & \multirow{4}{*}{-} & 80.55              \\
			\multicolumn{1}{c|}{}                          & charCBOW     & 54.91              & 72.96              & 58.40              &                    & 80.77              \\
			\multicolumn{1}{c|}{}                          & charSkipGram & 54.65              & 71.88              & 65.27              &                    & 80.31              \\
			\multicolumn{1}{c|}{}                          & DISA(T+P)    & \textbf{58.15}     & \textbf{77.51}     & \textbf{65.45}     &                    & \textbf{88.70}     \\ \hline
			\multicolumn{1}{c|}{\multirow{4}{*}{Review-5}} & Hsentic      & 58.44              & 74.73              & 69.08              & 83.15              & \multirow{4}{*}{-} \\
			\multicolumn{1}{c|}{}                          & charCBOW     & 56.73              & 72.47              & 57.06              & 85.77              &                    \\
			\multicolumn{1}{c|}{}                          & charSkipGram & 56.44              & 75.32              & 66.77              & 83.67              &                    \\
			\multicolumn{1}{c|}{}                          & DISA(T+P)   & \textbf{62.06}     & \textbf{85.65}     & \textbf{69.09}     & \textbf{88.85}     &                    \\ \hline
		\end{tabular}
	}
	
\end{table}

\subsection{Cross-domain Evaluation}
In this section, we examine how our model performs across different domains and datasets in order to validate the generalizability of our proposed method. Particularly for our model, we firstly pretrain the LSTM critic network on the training set. Then we fix the parameters of critic network and train the policy network on the same training set. Next, we co-train the LSTM critic network and policy network for 30 epochs. For other baselines, an LSTM network is trained using the same training set. By the end of each epoch, the development set of this training dataset and the other four datasets are tested. The epoch results are recorded. In the end, the testing result of the epoch which has the best development result are reported. The final results of the state-of-the-art methods are shown in Table~\ref{cross-tab}. 

\begin{figure}
	\centering
	\resizebox{0.4\textwidth}{!}{%
		\includegraphics[]{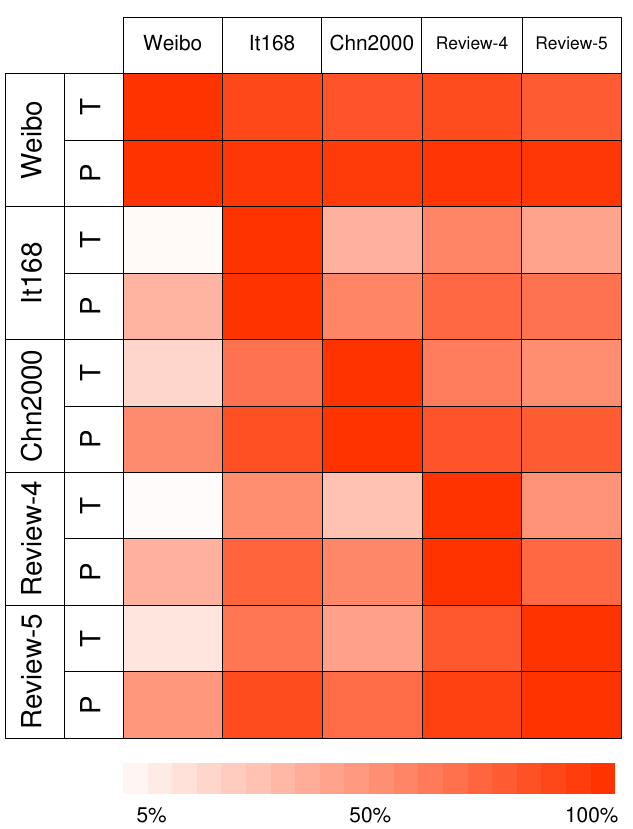} 
	}
	\caption{The proportion of tokens in testing sets that also appear in training sets. Rows are training sets(T denotes the textual token and P denotes the phonetic token) Columns are testing sets.}
	\label{overlap}       
\end{figure}

Results show that all methods reduce their performance compared to single dataset experiments due to the internal diversity of different dataset. Even though, our method still perform better than other baselines by an average of 6.50\% in accuracy. In addition to absolute performance, we also compute the average performance loss for each method across different datasets between single dataset case and cross-dataset case. It shows that our method has the least performance drop, which is 14.25\%. The performance drop for Hsentic, charCBOW and charSkipGram methods are 16.09\%, 15.69\%, 17.16\% respectively. We think it might be ascribed to that the proportion of shared phonetic tokens among datasets is larger than the portion of shared textual characters. Thus, phonetic features will have better transferability than textual features. Fig.~\ref{overlap} illustrates the proportion of common phonetic tokens as well as common textual tokens between each pair of datasets. The result in the figure agrees with our initial analysis.

\begin{figure*}
	\centering
	
	\begin{subfigure}{.5\textwidth}
		\centering
		
		\includegraphics[width=\linewidth]{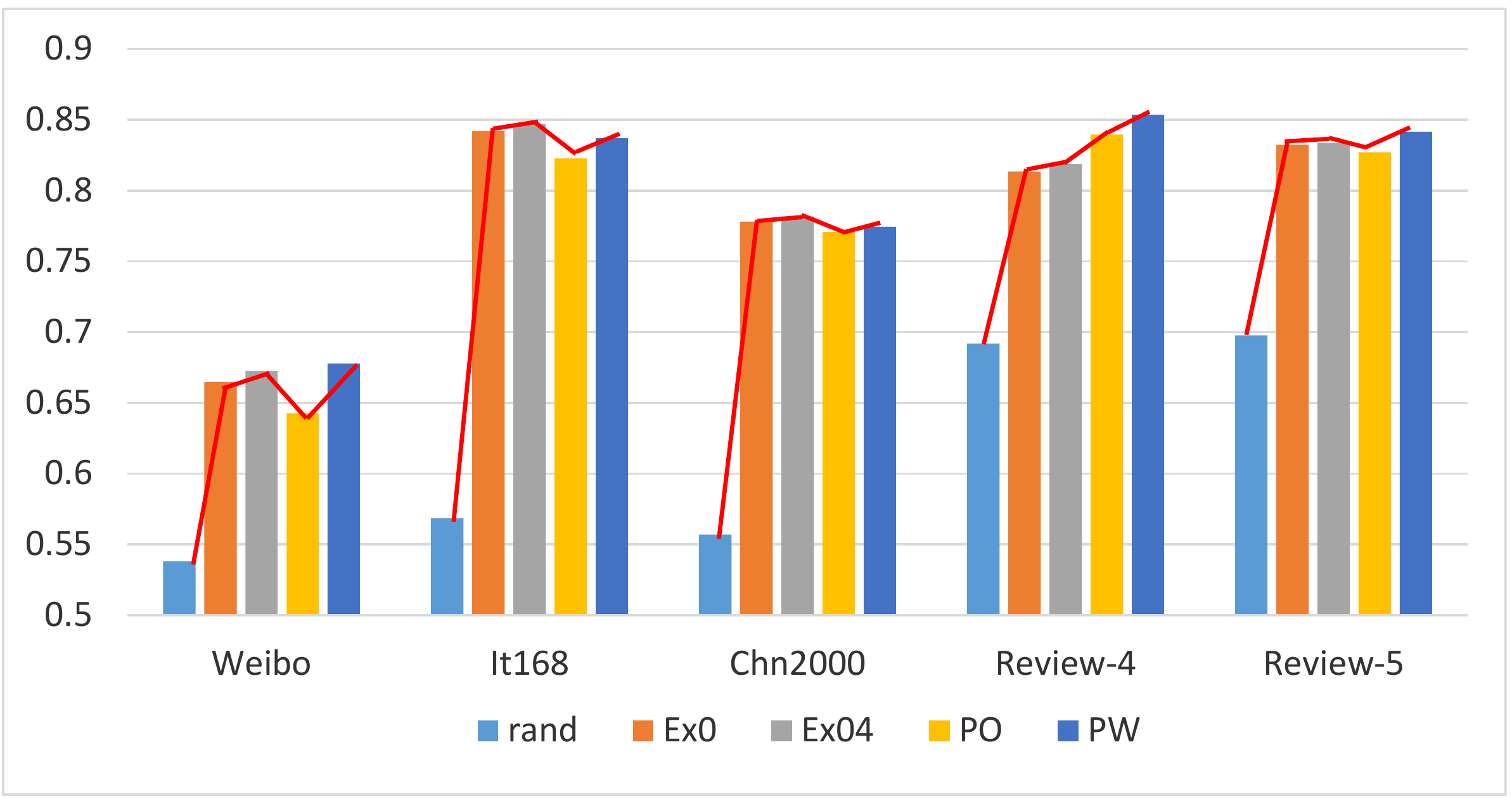}
	\end{subfigure}%
	\begin{subfigure}{.5\textwidth}
		\centering
		
		\includegraphics[width=\linewidth]{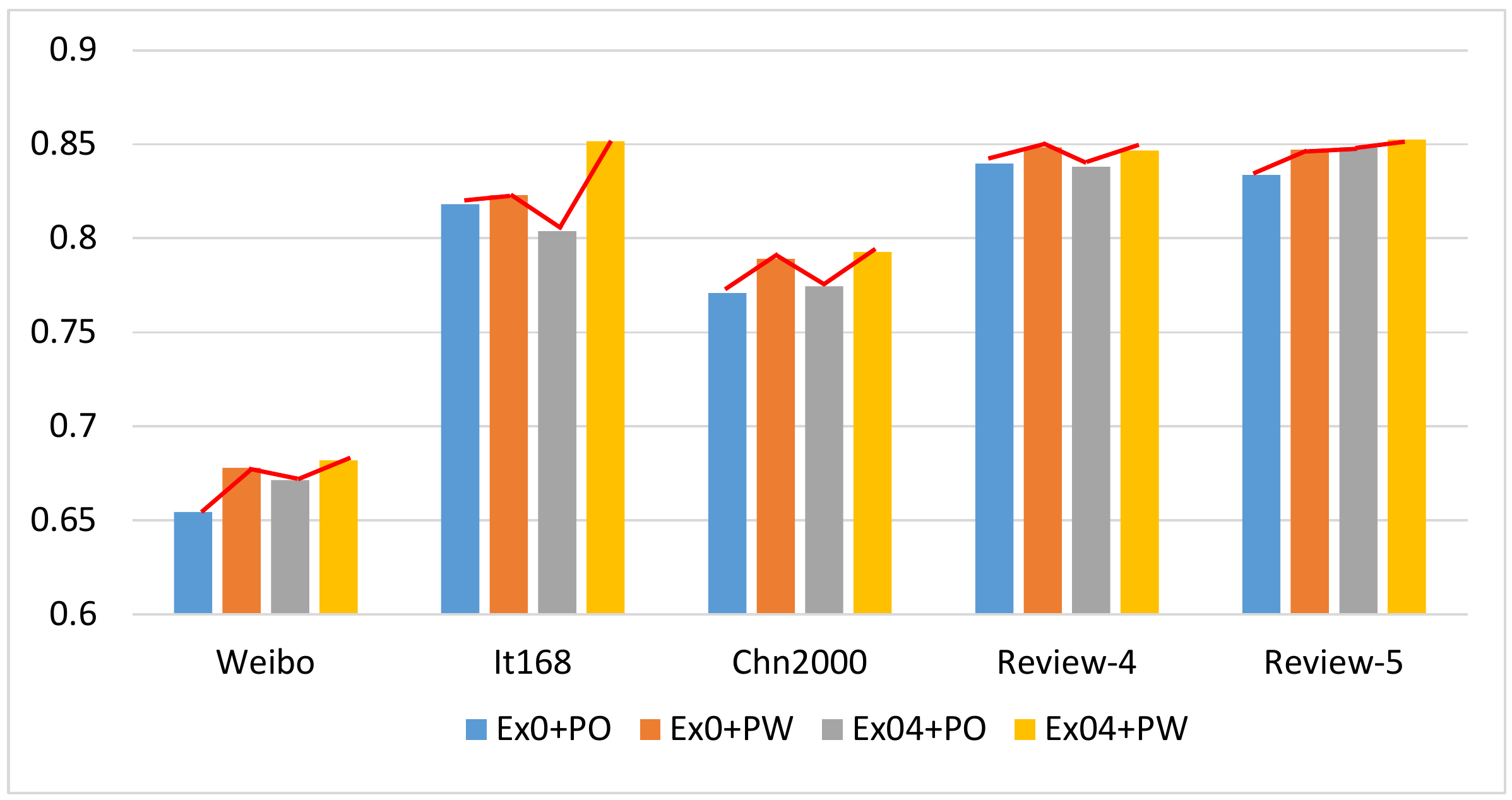}
	\end{subfigure}

	\caption{Performance comparison between phonetic ablation test groups. Rand denotes random generated embeddings. Ex0/Ex04 represent Ex embeddings without/with tones. The same is for PO/PW. + denotes a concatenation operation.}
	\label{figabl}
\end{figure*}

\subsection{Ablation Tests}
We conduct ablation tests in two steps: validating phonetic features and integration of phonetic features. The first step validates the contribution of phonetic features. The second step examines which specific combination of phonetic features works the best.

\subsubsection{Validating phonetic feature}
So far, we have examined the effectiveness of our model as a whole by comparing it with different baselines. In this section, we break down the proposed methods into a reinforcement learning framework and a set of features. First of all, we would like to validate if the performance gain mainly results from the reinforcement learning framework. To this end, we replace the phonetic features with random features. In particular, we generate random real-valued vectors as random phonetic feature for each character. Each dimension of the random phonetic feature vector is a float number between -1 to 1 sampled from a Gaussian distribution. Then, we use this random feature vector to represent each Chinese character and yielded the results in Table~\ref{rand-tab}.

In the comparison between the learned phonetic feature and random phonetic feature, we can observe that the learned feature outperforms random feature with at least 13\% in all datasets. This result indicates that the improvement of performance is due to the contribution of learned phonetic feature but not the training of classifiers. Phonetic feature itself is the cause and similar performance will not be achieved just by introducing random features. 

We plot the results in Fig.~\ref{figabl} on the left to amplify the difference. Moreover, we find that, whether extracted from audio clips or learned from pinyin corpus, phonetic features that contain intonation (Ex04 and PW) perform better than those without intonation (EX0 and PO) in all our experiments. 

This proves our initial argument that intonation plays an important role in representing Chinese sentiment. Nevertheless, we also discover that the performances of various learned phonetic features are not persistent. PW prevails in three datasets while Ex04 wins in the other two datasets. As the best two phonetic features are either extracted from audio clips or learned from pinyin corpus, it is expected to take the advantage of both sides. Thus we propose the ablation test of phonetic feature in different combination.

\begin{table}[]
	\centering
	\caption{Performance comparison between learned and random generated phonetic feature.}
	\label{rand-tab}
	\resizebox{0.5\textwidth}{!}{%
\begin{tabular}{ccccccc}
	\hline
	\multicolumn{2}{c}{}                                                                                                                   & Weibo          & It168          & Chn2000        & Review-4       & Review-5       \\ \hline
	\multicolumn{2}{c|}{Random phonetic feature (rand)}                                                                                    & 53.83          & 56.85          & 55.71          & 69.20          & 69.77          \\ \hline
	\multicolumn{1}{c|}{\multirow{4}{*}{\begin{tabular}[c]{@{}c@{}}Learned\\ phonetic\\ feature\end{tabular}}} & \multicolumn{1}{c|}{Ex0}  & 66.49          & 84.21          & 77.82          & 81.36          & 83.24          \\
	\multicolumn{1}{c|}{}                                                                                      & \multicolumn{1}{c|}{Ex04} & 67.28          & \textbf{84.69} & \textbf{78.18} & 81.88          & 83.38          \\
	\multicolumn{1}{c|}{}                                                                                      & \multicolumn{1}{c|}{PO}   & 64.28          & 82.30          & 77.09          & 83.97          & 82.71          \\
	\multicolumn{1}{c|}{}                                                                                      & \multicolumn{1}{c|}{PW}   & \textbf{67.80} & 83.73          & 77.45          & \textbf{85.37} & \textbf{84.18} \\ \hline
\end{tabular}
	}
\end{table}

\begin{table}[]
	\centering
	\caption{Performance comparison between different combinations of phonetic features}
	\label{phfuse-tab}
	\resizebox{0.5\textwidth}{!}{%
\begin{tabular}{cccccc}
	\hline
	& Weibo          & It168          & Chn2000        & Review-4       & Review-5       \\ \hline
	\multicolumn{1}{c|}{Ex0}     & 66.49          & 84.21          & 77.82          & 81.36          & 83.24          \\
	\multicolumn{1}{c|}{Ex04}    & 67.28          & 84.69          & 78.18          & 81.88          & 83.38          \\
	\multicolumn{1}{c|}{PO}      & 64.28          & 82.30          & 77.09          & 83.97          & 82.71          \\
	\multicolumn{1}{c|}{PW}      & 67.80          & 83.73          & 77.45          & \textbf{85.37} & 84.18          \\ \hline
	\multicolumn{1}{c|}{Ex0+PO}  & 65.45          & 81.82          & 77.09          & 83.98          & 83.38          \\
	\multicolumn{1}{c|}{Ex0+PW}  & 67.80          & 82.30          & 78.91          & 84.84          & 84.71          \\
	\multicolumn{1}{c|}{Ex04+PO} & 67.14          & 80.38          & 77.45          & 83.80          & 84.84          \\
	\multicolumn{1}{c|}{Ex04+PW} & \textbf{68.19} & \textbf{85.17} & \textbf{79.27} & 84.67          & \textbf{85.24} \\ \hline
\end{tabular}
	}
\end{table}

\begin{figure*}
	\centering
	\begin{subfigure}[b]{0.475\textwidth}
		\centering
		\includegraphics[width=\textwidth]{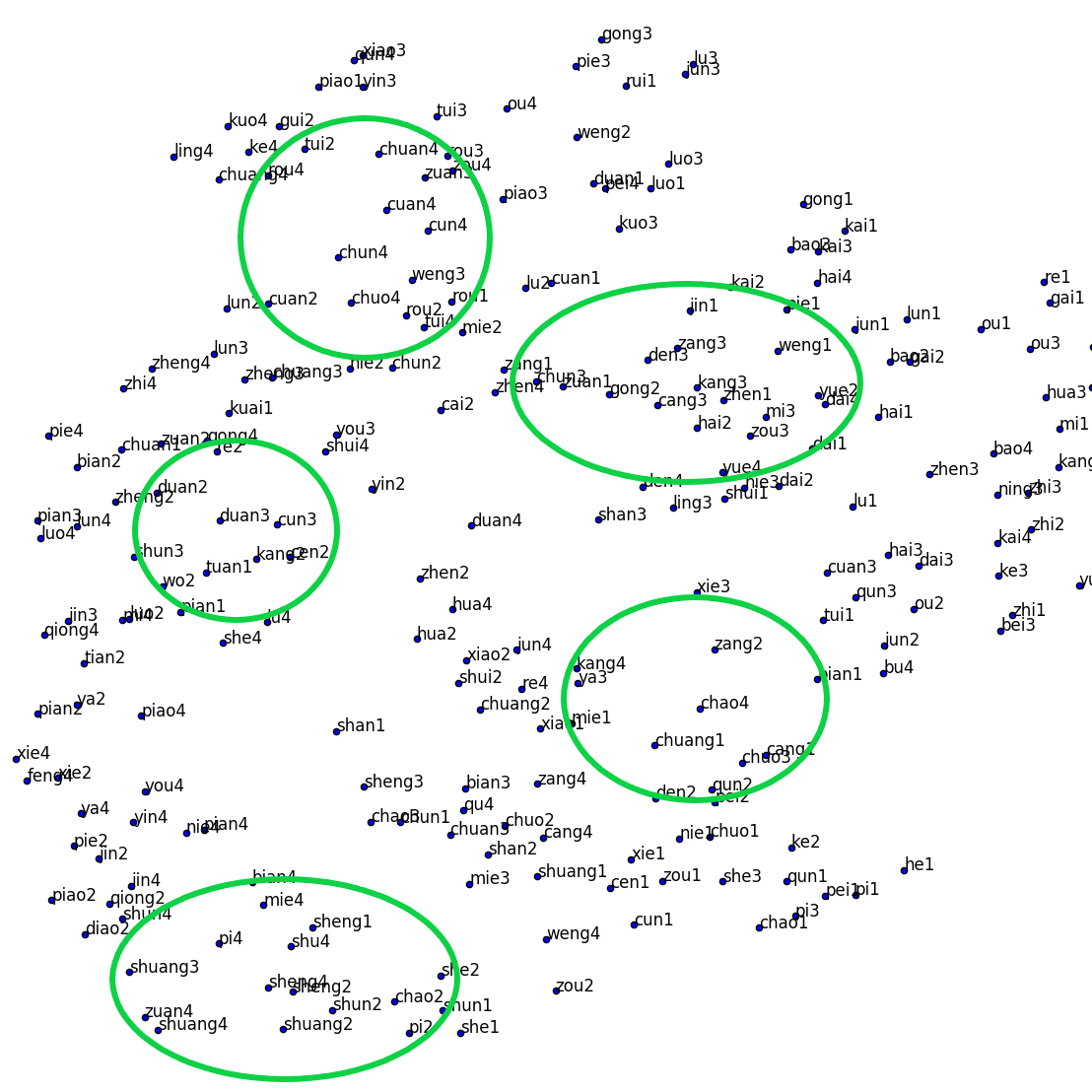}
		\caption[Network1]%
		{{\small Phonetic embedding Ex04}}    
		\label{figex04}
	\end{subfigure}
	\hfill
	\begin{subfigure}[b]{0.475\textwidth}  
		\centering 
		\includegraphics[width=\textwidth]{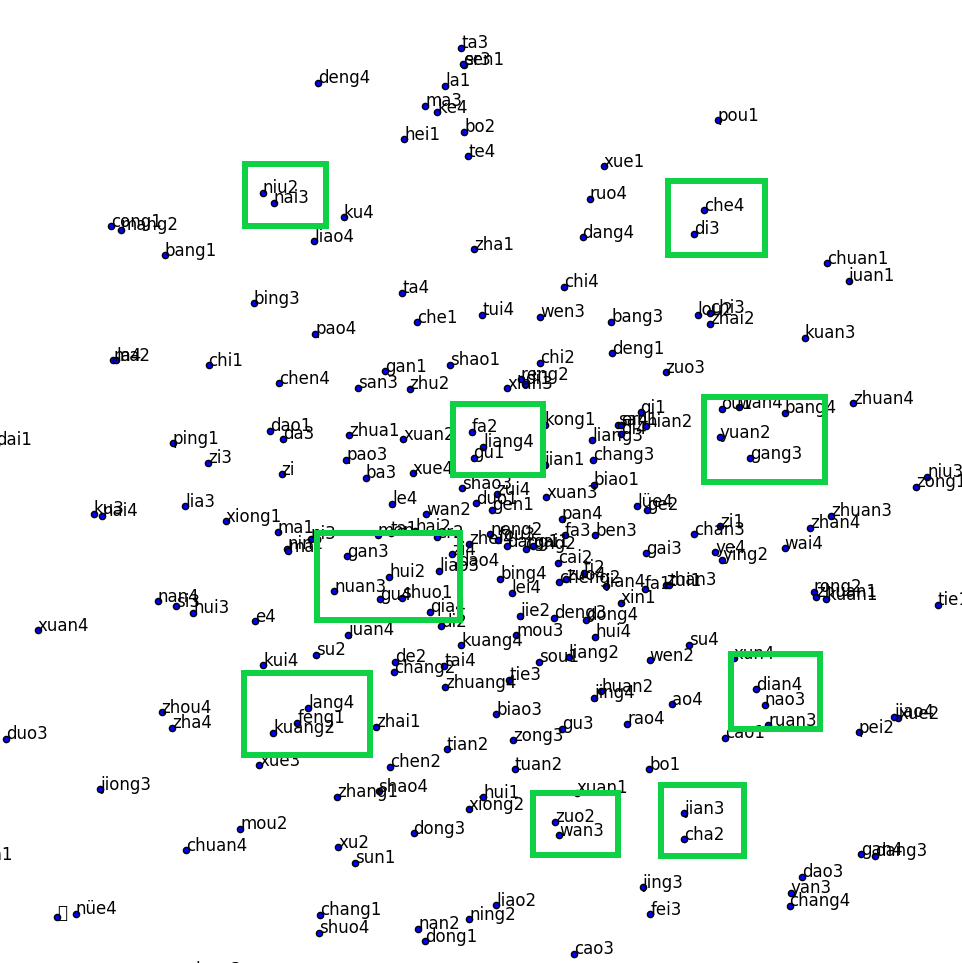}
		\caption[]%
		{{\small Phonetic embedding PW}}    
		\label{figpw}
	\end{subfigure}
	\vskip\baselineskip
	\begin{subfigure}[b]{0.475\textwidth}   
		\centering 
		\includegraphics[width=\textwidth]{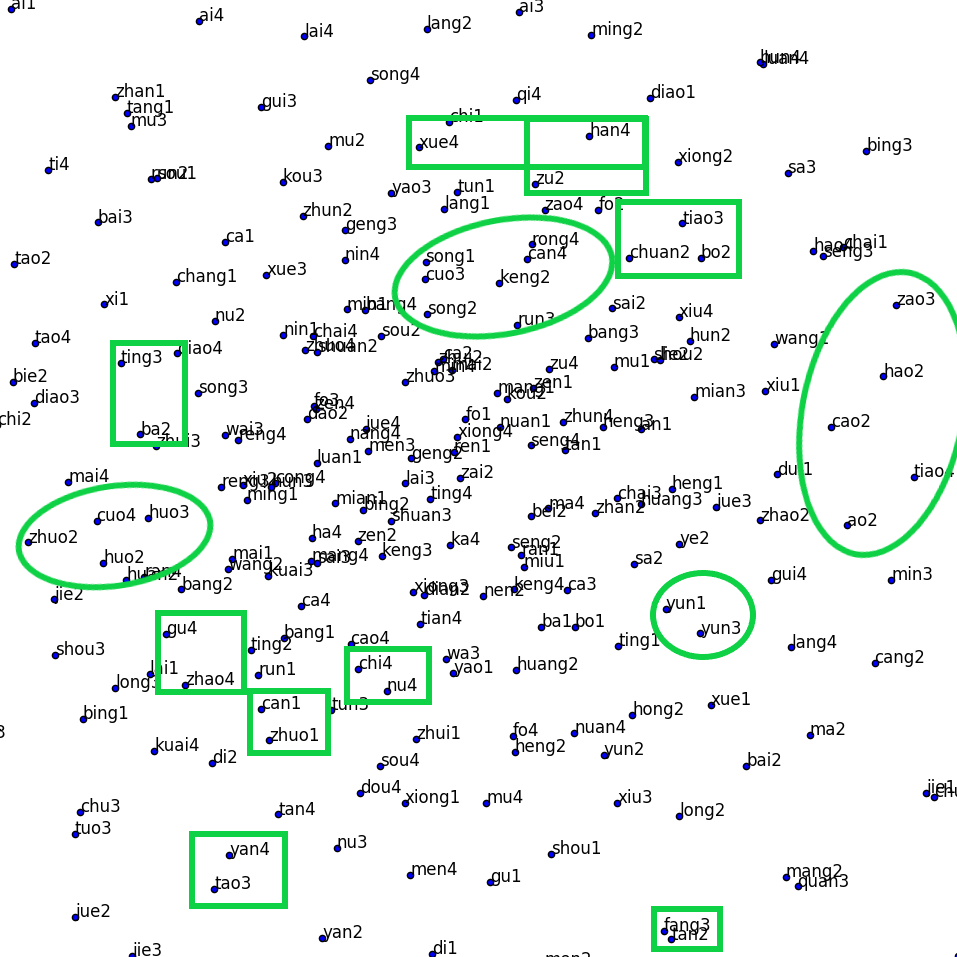}
		\caption[]%
		{{\small Phonetic embedding Ex04+PW (P)}}    
		\label{figex04pw}
	\end{subfigure}
	\quad
	\begin{subfigure}[b]{0.475\textwidth}   
		\centering 
		\includegraphics[width=\textwidth]{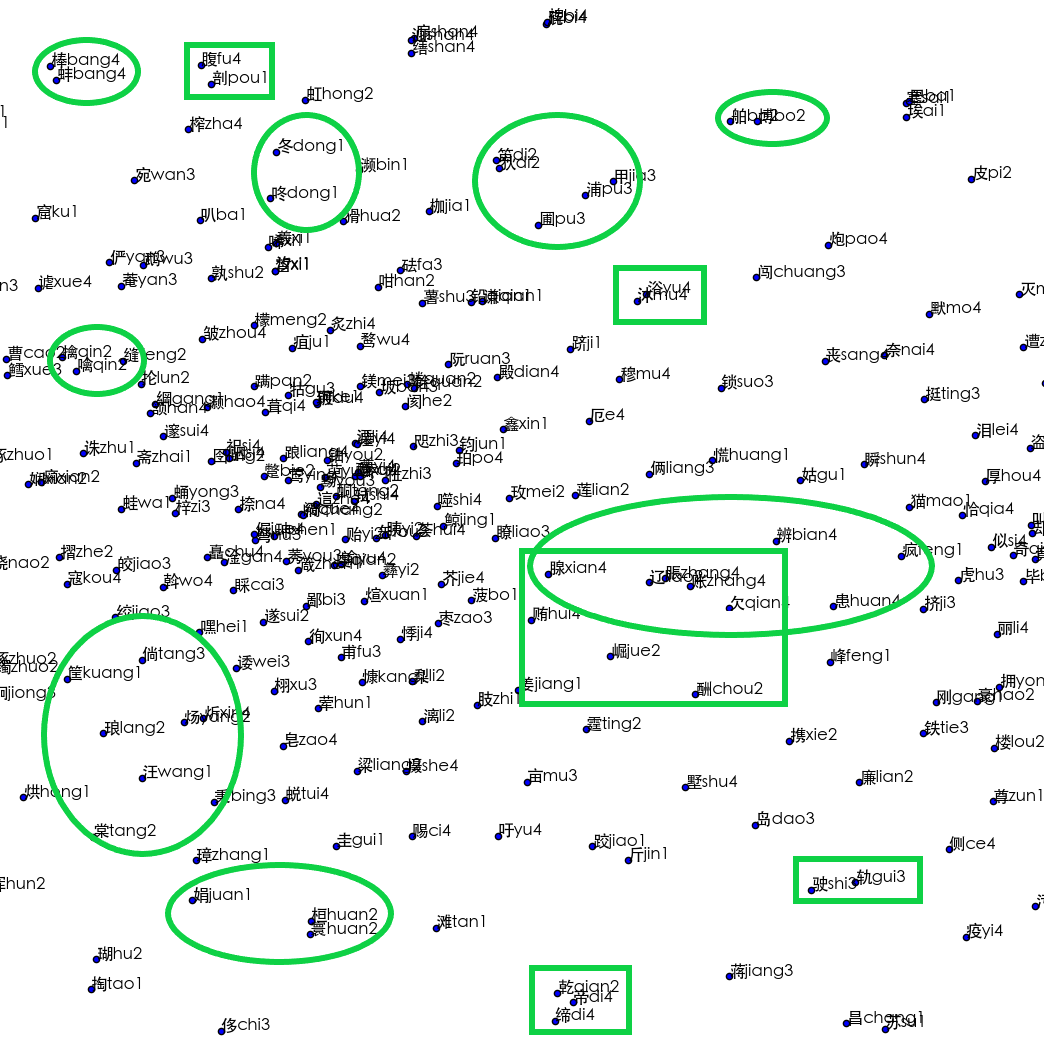}
		\caption[]%
		{{\small Phonetic embedding T+P}}    
		\label{figtex04pw}
	\end{subfigure}
	\caption[  ]
	{\small Selected t-SNE visualization of four kinds of phonetic-related embeddings. Circles cluster phonetic similarity. Squares cluster semantic similarity.} 
	\label{figclust}
\end{figure*}

\subsubsection{Integration of Phonetic Features}
We combine both extracted phonetic features and learned phonetic features to form four variations. The results are shown in Table~\ref{phfuse-tab} and plotted in Fig.~\ref{figabl} on the right.

As expected, the combination of Ex04 and PW prevails in four datasets and performs close to the best in the remaining dataset. Specifically, when we compare Ex04+PW with Ex04, there is an average improvement of 1.43\% across datasets. We believe the improvement was due to the semantic information provided by PW feature, as it was trained on the pinyin corpus. Contextual relation was designed to be encoded in embeddings. By merging embedding features to extracted features, the combination feature would also encode certain semantics, which we would show in the following section. Correspondingly, if we compare Ex04+PW with PW, the performance improvement was 0.80\% on average. 

This would be explainable due to Ex04 features extracted information that can only be conveyed in pronunciation. As we introduced in the start, the deep phonemic orthography has enabled Chinese pronunciation to encode meanings that were not represented in the text. The English text, in contrast, originally was designed to mimic pronunciation~\cite{albrow1972english}. Due to the heterogeneity between textual and phonetic representation of the Chinese language, it is reasonable to unveil the magic behind Chinese phonetics. In summary, we have shown that both the intonation variation and deep phonemic orthography contributed to Chinese sentiment analysis task.

\subsection{Visualization}
In this section, we visualize four kinds of phonetic-related embeddings. The are Ex04, PW, Ex04+PW (P) and T+P.

As shown in Fig.~\ref{figex04}, pinyins that have similar pronunciations (vowels) are close to each other in the embedding space. This observation matches our experimental purpose that the Ex04 feature will encode phonetic information (such as similarity) among different pronunciations. Secondly, as can be seen in Fig.~\ref{figpw}, we visualize the embeddings of PW. Since it was learned on the phonetic corpus, certain semantics are expected to be encoded. In reality, we do find semantic closeness in the visualization. The squares are some examples we spotted. For instance, `Niu2' and `Nai3' are together due to `Niu2 Nai3(milk)'. `Dian4' and `Nao3' are together due to `Dian4 Nao3 (computer)'. `Jian3' and `Cha2' are together due to `Jian3 Cha2 (inspection)'. Next, we visualize the combined embedding, Ex04+PW, which is also the main phonetic feature we use in our model in Fig.~\ref{figex04pw}. Unsurprisingly, we observe that this feature combines the characteristics both from Ex04 and PW, because this embedding clusters not only phonetic similarity but also semantic similarity. Finally, we visualize the fused embedding of T+P in Fig.~\ref{figtex04pw}. In addition to the characteristics displayed in Ex04+PW (P), the fused T+P appends with Chinese textual characters. For example, \begin{CJK*}{UTF8}{gbsn}沐\end{CJK*}Mu4 and \begin{CJK*}{UTF8}{gbsn}浴\end{CJK*}Yu4 stayed together because of semantics(bath). \begin{CJK*}{UTF8}{gbsn}桓\end{CJK*}Huan2 and \begin{CJK*}{UTF8}{gbsn}寰\end{CJK*}Huan2 stayed together because of phonetics. It can be concluded that the fused embeddings capture certain phonetic information from phonetic features and semantic information from textual embeddings.
This shows us why phonetic-enrich text representation could level up the performance in sentiment analysis compared with pure textual representation.

\section{Conclusion}\label{sec:Conclusion}

Modern Chinese pronunciation system (pinyin) provides a new perspective in addition to the writing system in representing Chinese language. Due to its deep phonemic orthography and intonation variations, it is expected to bring new contributions to the statistical representation of Chinese language, especially in the task of sentiment analysis. 

To the best of our knowledge, we are the first to present an approach to learn phonetic information out of pinyin (both from audio clips and pinyin token corpus) and design a network to disambiguate intonations. With the learned phonetic information, we integrate it with textual and visual features to create new Chinese representations. Experiments on five datasets demonstrated the positive contribution of phonetic information to Chinese sentiment analysis.

Even though our method only examines Chinese language, it suggests greater potential for languages that also carry the deep phonemic orthography characteristic, such as Arabic and Hebrew. In the future, we plan to extend the work in the following directions: firstly, we would like to explore better fusion methods to combine different modalities; secondly, we would like to study the word-level phonetic information.

\bibliographystyle{IEEEtran}
\bibliography{ref}

\end{document}